# An Algebraic Approach for the MIMO Control of Small Scale Helicopter

**A. Budiyono*** and **T. Sudiyanto**[†]

*Center for Unmanned System Studies
Institut Teknologi Bandung, Indonesia
e-mail: agus.budiyono@ae.itb.ac.id

[†] Aeronautics and Astronautics Department
Institut Teknologi Bandung, Indonesia

**Abstract**

The control of small-scale helicopter is a MIMO problem. To use of classical control approach to formally solve a MIMO problem, one needs to come up with multi-dimensional Root Locus diagram to tune the control parameters. The problem with the required dimension of the RL diagram for MIMO design has forced the design procedure of classical approach to be conducted in cascaded multi-loop SISO system starting from the innermost loop outward. To implement this control approach for a helicopter, a pitch and roll attitude control system is often subordinated to a, respectively, longitudinal and lateral velocity control system in a nested architecture. The requirement for this technique to work is that the inner attitude control loop must have a higher bandwidth than the outer velocity control loop which is not the case for high performance mini helicopter. To address the above problems, an algebraic design approach is proposed in this work. The designed control using s-CDM approach is demonstrated for hovering control of small-scale helicopter simultaneously subjected to plant parameter uncertainties and wind disturbances.

## 1 Introduction

The control for a small scale helicopter has been designed using various methods. During the period of 1990s, the classical control systems such as single-input-single-output SISO proportional-derivative (PD) feedback control systems have been used extensively. Their controller parameters were usually tuned empirically. This trial-and-error approach to design an "acceptable" control system however is not agreeable with complex multi-input multi-output MIMO systems with sophisticated performance criteria. For more advanced multivariable controller synthesis approaches, an accurate model of the dynamics is required. To control a model helicopter as a complex MIMO system, an approach that can synthesize a control algorithm to make the helicopter meet performance criteria while satisfying some physical constraints is required. More recent development in this area include the use of optimal control (Linear Quadratic Regulator) implemented on a small aerobatic helicopter designed at MIT [1]. Similar approach based on μ-synthesis has been also independently developed for a rotor unmanned aerial vehicle at UC Berkeley [2]. An adaptive high-bandwidth helicopter controller algorithm was synthesized at Georgia Tech. [3].

To address a MIMO problem, LQR and $H_\infty$ are the most popular control design procedures. However, these methods are not up to expectation for practical application in aerospace community, because of the following reasons [4];

1. Parameter tuning procedures are not provided
2. Weight selection rules are not established
3. The controller order is unnecessarily high
4. Robustness is guaranteed only for predefined ones
5. Some times, traditionally accepted good controllers are excluded
6. Extension of gain scheduling or inclusion of proper saturation of state variable is difficult
7. LQR and LQG designs sometimes fail to produce a robust controller for the plant with flexibility

Due to the above limitations, the classical control design by experienced engineers is still common in the aerospace community. However, the inheritance of such experiences is often difficult, thus an improvement of the method is highly desired. In particular, the drawbacks of this approach to be used as control design tool for a small scale helicopter can be elaborated as follows.

1. The control of small-scale helicopter is a MIMO problem. To use of classical control approach to formally solve a MIMO problem, one needs to come up with multi-dimensional Root Locus diagram to tune the control parameters. Such diagram however is not presently available
2. The problem with the required dimension of the RL diagram for MIMO design has forced the design procedure of classical approach to be conducted in cascaded multi-loop SISO system starting from the innermost loop outward. As shown in the design example in this work, this type of approach is unnecessarily cumbersome
3. The cascaded multi-loop SISO approach has limitations in its implementation. To implement this control approach for a helicopter, a pitch and roll attitude control system is often subordinated to a, respectively, longitudinal and lateral velocity control system in a nested architecture. The requirement for this technique to work is that the inner attitude control loop must have a higher bandwidth than the outer velocity control loop. For a class of high-performance helicopters, such as the X-Cell 60, or helicopters where this bandwidth separation is not sufficient, a simultaneous design is necessary[5].
4. The classical control approach is associated with the use of transfer function which can become inaccurate when pole-zero cancellation occurs due to uncontrollable and unobservable modes

To address the above problems, a third approach generally called as algebraic design approach is proposed in this paper.

## 2  Dynamics of small scale helicopter

The dynamics model of a small scale RUAV has been elaborated in [6] for X-Cell 60 RC helicopter. The model is developed using first principle approach. The mathematical model was developed using basic helicopter theory accounting for particular characteristic of the miniature helicopter. Most of the parameters were measured directly, several were estimated using collected data from simple flight test experiment involving step and impulse response in various actuator inputs. No formal system identification procedures are required for the proposed model structure.

### 2.1  First principle approach Length
Beyond the previous work in [1], the calculation of stability and control derivatives to construct the linear model is presented in detail. The analytical model derivation allows the comprehensive analysis of relative dominance of vehicle states and input variables to force and moment components. And hence it facilitates the development of minimum complexity small scale helicopter dynamics model that differs from that of its full-scale counterpart. In the presented simplified model, the engine drive-train dynamics and inflow dynamics are not necessary to be taken into consideration. The additional rotor degrees of freedom for coning and lead-lag can be omitted for small scale helicopters. It is demonstrated analytically that the dynamics of small scale helicopter is dominated by the strong moments produced by the highly rigid rotor. The dominant rotor forces and moments largely overshadow the effects of complex interactions between the rotor wake and fuselage or tail. This tendency substantially reduces the need for complicated models of second-degree effects typically found in the literature on full-scale helicopters. .

### 2.2  Linear model of small scale helicopter
The presented approach is not limited to specific trim conditions like hover or forward flights and therefore can be used to develop a global model of small scale rotorcraft vehicle to the purpose of practical control design.

The developed model is presented in the form of state-space with ten states and four inputs. Subsequently, it was shown by the Frobenius' norm analysis that the coupling between longitudinal and lateral directional is small. Thus the control design uses the decoupled model in longitudinal and lateral directional mode [6].

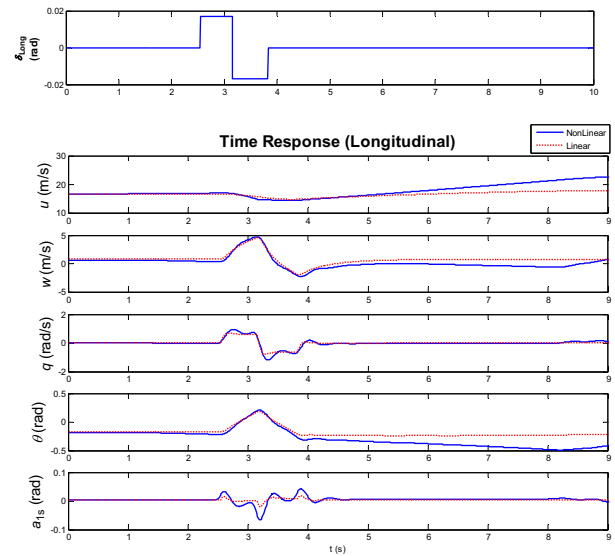

**Figure 1 Response comparison between linear and nonlinear model**

## 3  Coefficient Diagram Method for Control Design

In this study, a novel approach pioneered by Manabe [7] using algebraic representation applied to polynomial loop in the parameter space, is proposed as control design candidate. With this technique, a unique coefficient diagram is used as the means to convey the necessary practical design information and as the criteria of good design. The eventual application of CDM in this work is in the LQ design framework to be elaborated in the next section.

### 3.1  Design principle
**Mathematical model.** The mathematical model of the CDM design is described in general as a block diagram shown in **Figure 2**. In this figure, $r$ is the reference input signal, $u$ is the control signal, $d$ is the disturbance and $n$ is the noise generated by the measuring device at the output; N(s) and D(s) are the numerator and denominator polynomial of the plant transfer function, respectively. A(s), F(s) and B(s) are the polynomials associated with the CDM controller which are the denominator polynomial matrix of the controller, the reference and the feedback numerator polynomial matrix of the controller respectively. For MIMO case, the variables and components are in the form of vectors and matrices with the appropriate dimension.

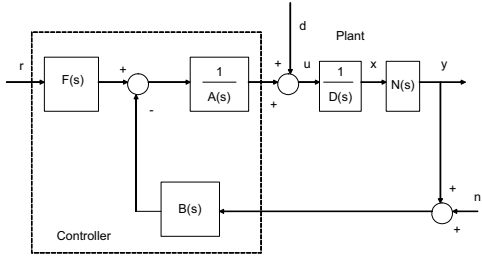

Figure 2 CDM block diagram

The plant equation is given by:

$$y = N(s)x$$
$$y = \frac{N(s)}{D(s)}(u+d) \quad (1)$$

which after some algebraic manipulation, can be completely written as:

$$y = \frac{N(s)F(s)}{P(s)}r + \frac{A(s)N(s)}{P(s)}d - \frac{N(s)B(s)}{P(s)}n \quad (2)$$

where P(s) is the closed-loop system polynomial matrix expressed by:

$$P(s) = A(s)D(s) + B(s)N(s) = \sum_{i=0}^{n} a_i s^i \quad (3)$$

The characteristic polynomial $\Delta(s)$ is given by:

$$\Delta(s) = \det P(s) \quad (4)$$

To write the input-output relation of the system, the expression for the state and the controllers are needed. The controller equation can be written as:

$$A(s)u = F(s)r - B(s)(n+y) \quad (5)$$

Whereas the state equation can be obtained by eliminating u and y from the controller and output equations as follows:

$$P(s)x = F(s)r + A(s)d - B(s)n \quad (6)$$

Combining the output, state and controller equations, Eqs.(1), (5) and (6), the matrix input-output equation can finally be expressed as:

$$\begin{bmatrix} x \\ y \\ z \end{bmatrix} = \frac{1}{\Delta(s)} \begin{bmatrix} I \\ N(s) \\ D(s) \end{bmatrix} adj A(s)[F(s)r + A(s)d - B(s)n] - \begin{bmatrix} 0 \\ 0 \\ d \end{bmatrix}$$
(7)

**CDM controller design.** The design parameters in CDM are the stability indices $\gamma_i's$, the stability limit indices $\gamma_i^{*'}s$ and the equivalent time constant, $\tau$. The stability index and the stability limit index determine the system stability and the transient behavior of the time domain response. In addition, they determine the robustness of the system to parameter variations. The equivalent time constant, which is closely related to the bandwidth, determines the rapidity of the time response. Those parameters are defined as follows:

$$\gamma_i = \frac{a_i^2}{(a_{i+1}a_{i-1})}, \quad i = 1, 2, \cdots, n-1$$
$$\tau = \frac{a_1}{a_0}$$
$$\gamma_i^* = \frac{1}{\gamma_{i+1}} + \frac{1}{\gamma_{i-1}}, \quad \gamma_0 = \gamma_n \triangleq \infty$$
(8)

where $a_i$'s are coefficients of the characteristic polynomial $\Delta(s)$. The equivalent time constant of the i-th order $\tau_i$ is defined in the same way as $\tau$.

$$\tau_i = \frac{a_{i+1}}{a_i} \quad (9)$$

By using the above equations, the relation between $\tau_i$'s can be written as:

$$\frac{\tau_i}{\tau_{i-1}} = \frac{a_{i+1}}{a_i}\frac{a_{i-1}}{a_i} = \frac{1}{\gamma_i} \quad (10)$$

Also, by simple manipulation, $a_i$ can be written as:

$$a_i = \tau_{i-1} \cdots \tau_1 \tau a_0$$

$$a_i = \frac{a_0 \tau^i}{\gamma_{i-1}\gamma^2_{i-2}\cdots\gamma^{i-2}_{2}\gamma^{i-1}_{1}}, \quad i \geq 2 \quad (11)$$

The characteristic polynomial can then be expressed as:

$$\Delta(s) = a_0 \left[ \left\{ \sum_{i-2}^{n} \left( \prod_{j=1}^{i-1} \frac{1}{\gamma^j_{i-j}} \right)(\tau s)^i \right\} + \tau s + 1 \right] \quad (12)$$

The sufficient condition for stability is given as:

$$a_i > 1.12 \left[ \frac{a_{i-1}}{a_{i+1}} a_{i+2} + \frac{a_{i+1}}{a_{i-1}} a_{i-2} \right]$$

$$\gamma_i > 1.12\gamma_i^*, \forall i = 2,3,\cdots,n-2 \quad (13)$$

And the sufficient condition for instability is:

$$a_{i+1}a_i \leq a_{i+2}a_{i-1}$$

$$\gamma_{i+1}\gamma_i \leq 1, \quad \text{for some } i = 1,\cdots,n-2 \quad (14)$$

### 3.2 Application to helicopter control

To control a model helicopter as a complex MIMO system, an approach that can synthesize a control algorithm to make the helicopter meet performance criteria while satisfying some physical constraints is required. Overall it is always desired to have a controller that can accommodate the unmodeled dynamics or parameter changes and perform well in such situations. Coefficient Diagram Method (CDM) is chosen as the candidate to synthesize such a controller due to its simplicity and convenience in demonstrating integrated performance measures including equivalent time constant, stability indices and robustness. To use CDM approach, the dynamics model should be first developed. In our case, the dynamics model of the small scale helicopter has been derived analytically.

### 4  Hover and Cruise Control Design

To demonstrate the viability of the algebraic approach, the hover stabilization and cruise control are taken as case studies. The well-known hover control problem represents a unique challenging problem in the real world. Many helicopter applications (both manned and unmanned) require the stable hovering capability for different missions: video air surveillance, air photography, precision targeting etc. The preliminary study conducted by the author for R-50 Yamaha helicopter hover control was given in [8]. The control during cruise is also important for different types of missions and serves as the basis for autonomous capability such as way-point following navigation and auto-piloting. As the baseline control design, multi-loop SISO system based on classical approach is first proposed. The CDM method is then proposed as an improvement of such an approach. For the sake of brevity, the speed control is taken as an example.

### 4.1 Classical approach to speed control

The classical approach to speed control of small scale helicopter is the extension of the SAS and Hold system in a cascaded control architecture. For the purpose of illustration, the result is presented for the design of forward speed $u$ control. Figure 3 describes the root locus diagram of speed control to be used for control parameter optimization. The gain selected for the design using the root locus diagram is ku = -0.0221. The time response diagram for the speed control subject to step is given in Figure 4.

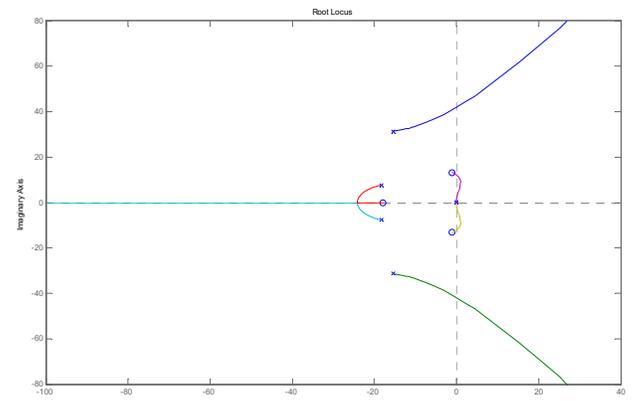

Figure 3 Root locus speed control system

The time response diagram shows that the design for the forward speed control using classical approach does not meet the expectation. The multi-loop cascaded design is also in general cumbersome and ineffective to handle the MIMO problem.

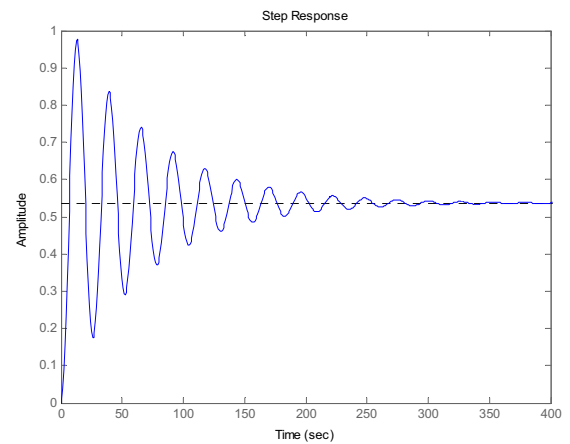

Figure 4 Step response speed control system

**4.2 Coefficient Diagram Method for speed**

As design example using CDM, the following section elaborates the procedure for developing *u* control and *v* control.

For denumerator and numerator polynomials in the pitch cyclic channel for the longitudinal vertical model are calculated as the following:

$$\Delta(s) = s^5 + 31.65s^4 + 321.7s^3 + 41.4s^2 + 11.02s + 0.9$$
$$(u \to \delta_{lon})(s) = -41.8s^4 - 840s^3 - 8919s^2 - 131290s - 12522$$
$$(q \to \delta_{lon})(s) = 904s^4 + 13457s^3 + 1710s^2 + 4s$$
$$(\theta \to \delta_{lon})(s) = 901s^3 + 13417s^2 + 1705s + 4$$
$$(w \to \delta_{lon})(s) = 0.67s^3 + 14.5s^2 + 214s + 6.85 \quad (15)$$

The corresponding coefficient diagrams for the above numerator and denominator polynomials are given in the following figures. The control design objective in this case is to change the coefficient of polynomial in order to stabilize the system using the appropriate feedback. By observation, it is more effective to use $\delta_{lon}$ to achieve the objective since $\delta_{col}$ is effective only when the vertical velocity feedback is used.

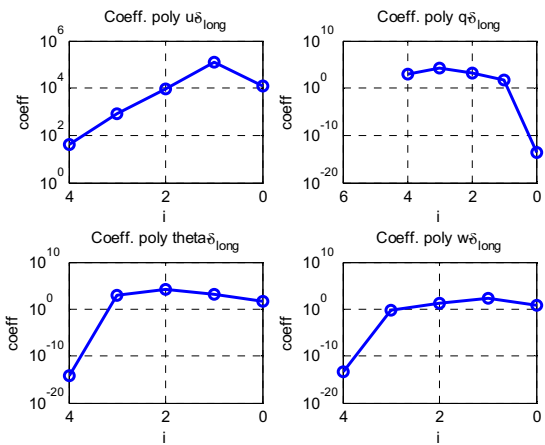

Figure 5 Coefficient diagrams for numerator poly - $\delta_{lon}$

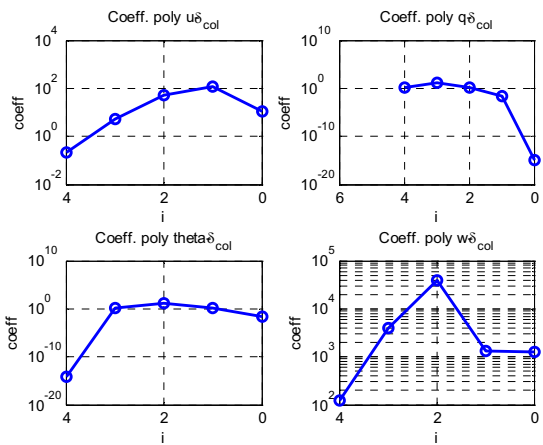

Figure 6 Coefficient diagrams for numerator polynomials - $\delta_{col}$

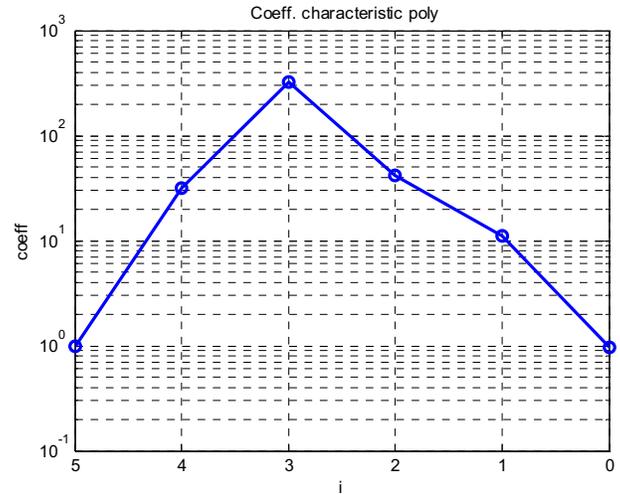

Figure 7 Coefficient diagram for denumerator polynomials -*u* control

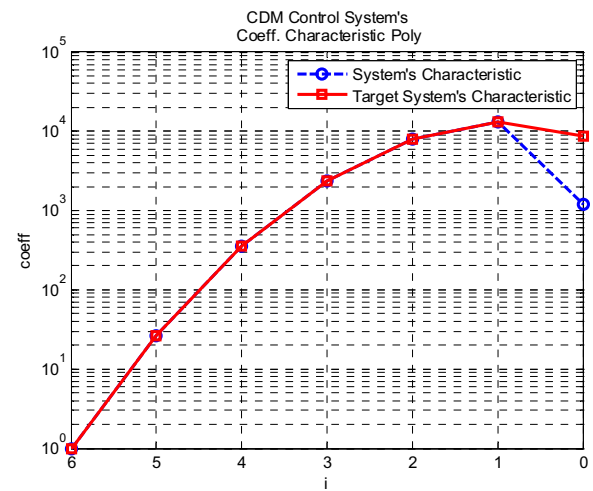

Figure 8 Coefficient diagram for denumerator polynomials -*u* control

Using the CDM diagram, it can be observed that we can choose the PID controller such that:

$$s\delta_{lon} = k_0 u_r - [(k_0 + k_1 s)u + (k_2 + k_3 s)\theta + k4w] \quad (16)$$

The new characteristic polynomial P(s) then becomes:

$$P(s) = s\Delta(s) + (k_0 + k_1 s)(u \to \delta_{lon}) +$$
$$(k_2 + k_3 s)(\theta \to \delta_{lon}) + k4(w \to \delta_{lon})$$

$$P(s) = s^6 + a_5 s^5 + a_4 s^4 + a_3 s^3 + a_2 s^2 + a_1 s + a_0 \quad (17)$$

Solving the Diophantine equation,

$-41.80k1 + 31.65 = a5,$
$-41.80k0 - 840.09k1 + 901.2k3 + 321.74 + .676k4 = a4,$
$-840.09k0 + 13416.5k3 - 8918.95k1 + 901.27k2 + 41.42 + 14.53k4 = a3,$
$11.020 + 214.61k4 + 1705.16k3 - 131287.6k1 + 13416.5k2 - 8918.9k0 = a2,$
$40.846k3 + .9583 - 131287.6k0 - 12522.15k1 + 1705.16k2 + 6.85k4 = a1$
(18)

we get the value for the gains as the following:
$$\begin{aligned} k0 &= -0.09194694469843 \\ k1 &= 0.11932152086877 \\ k2 &= 1.46173961265554 \\ k3 &= 0.13434584413061 \\ k4 &= 13.48966455977404 \end{aligned}$$ (19)

A similar procedure is implemented for the design of side velocity control. The result of the design is demonstrated in the following figures, including the test for the robustness due to an impulsive disturbance. The responses of a unit doublet in the input and to an impulse disturbance at $t=35$s are given. The controllers design using CDM all show a good disturbance rejection with zero steady-state error. The figures also show the robustness of control due to modeling uncertainty. For the longitudinal case, the control is tested by allowing the stability derivatives $x_u, x_{a_{1s}}$ and $m_q$ to vary by $\pm 30\%$.

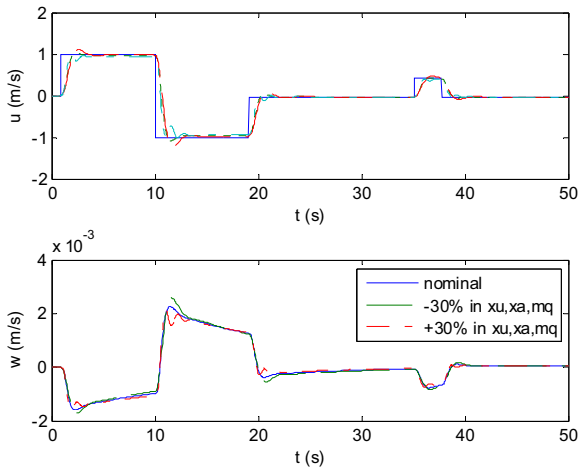

Figure 9 Forward speed control design with uncertainty in $x_u, x_{a_{1s}}$ and $m_q$

The superiority of CDM design over classical approach has been demonstrated through results comparison of speed control design. The result is further extended for the design of MIMO control system. The following section details the implementation of CDM in the optimal control framework, using the so-called squared-Coefficient Diagram Method (s-CDM). The proposed technique is an extension from previous work by Manabe [7] where the implementation of s-CDM is given for a simple SIMO problem.

## 5 Squared-Coefficient Diagram Method (s-CDM) for helicopter control

The motivation behind this approach is the limitation of the existing LQR/LQG techniques. The primary concern for LQR/LQG designs has been the analytical weight selection for such techniques. Numerous efforts have been given in the literature and the only workable solution to date is obtained through iteration. In addition to this problem, LQR and LQG sometimes fail to produce a robust controller for the plant with poles at the vicinity of the imaginary axis. As typical aerospace vehicles have this kind of behavior (i.e. the phugoid mode), the LQR or LQG should be used with caution. In what follows, s-CDM is proposed in conjunction with LQ design in which analytical weight selection is presented. The robustness of the control will be tested against parameter uncertainties and disturbances.

### 5.1 LQR Framework

The derivation is started by introducing the standard LQR formulation. In LQR framework, the plant is expressed in the state space expression:

$$\dot{\bar{x}} = \mathbf{A} \cdot \bar{x} + \mathbf{B} \cdot \bar{u}$$ (20)

where $\bar{x}$ is a vector of dimension $n$, and $\bar{u}$ is an input vector. LQR design is made to minimize the performance index $J$ given as

$$J = \int_0^\infty \left[ \bar{x}^T \cdot \mathbf{Q} \cdot \bar{x} + \bar{u}^T \cdot \mathbf{R} \cdot \bar{u} \right] dt$$ (21)

where $\mathbf{R}$ is positive definite, but $\mathbf{Q}$ is not necessarily sign definite. The first term represents the regulation or tracking performance and the second term the minimization of control power. The closed-loop poles of the system with the feedback control are given by the stable eigen values of the Hamiltonian $\mathbf{H}$, where no eigen values lie on the imaginary axis.

$$\mathbf{H} = \begin{bmatrix} \mathbf{A} & -\mathbf{R}^{-1}\mathbf{B}^T \\ -\mathbf{Q} & -\mathbf{A}^T \end{bmatrix}$$ (22)

When the characteristic polynomial is given as in

$$P_{(s)} = a_n s^n + a_{n-1} s^{n-1} + \cdots + a_1 s + a_0 = \sum_{i=0}^{n} a_i s^i$$ (23)

the following relation is obtained.

$$\frac{P_{(-s)}P_{(s)}}{a_n^2} = (-1)^n \det(sI_{2n} - H) \quad (24)$$

Therefore, if $P_{(s)}$ is designed by CDM such as done in the previous section, the weight $Q$ can be found. On the contrary, if $Q$ is specified and LQ design is made, $P_{(s)}$ can be obtained and it will be assessed in term of CDM.

**5.2 Squared polynomial in s-CDM**
For a given polynomial $P_{(s)}$, $P_{(-s)} P_{(s)}$ is a polynomial in $-s^2 = \Omega$, denoted as $PP_{(\Omega)}$. $PP_{(\Omega)}$ will be called the squared polynomial of $P_{(s)}$ hereafter, and $P_{(s)}$ will be called the original polynomial of $PP_{(\Omega)}$.

$$P_{(-s)}P_{(s)} = PP_{(-s^2)} = PP_{(\Omega)} \quad (25)$$

If $PP_{(\Omega)}$ has no positive real roots, there exists one original polynomial $P_{(s)}$ which is stable. This polynomial will be called the square-root polynomial of $PP_{(\Omega)}$. When $P_{(s)}$ is a characteristic polynomial, $P_{(s)}$ is the square-root polynomial of the squared polynomial $PP_{(\Omega)} = P_{(-s)} P_{(s)}$, because it is stable. The coefficients of these polynomials are selected as follows.

$$P_{(s)} = a_n s^n + a_{n-1} s^{n-1} + \cdots + a_1 s + a_0 = \sum_{i=0}^{n} a_i s^i \quad (26)$$

$$\begin{aligned}
PP_{(\Omega)} &= aq_n \Omega^n + aq_{n-1}\Omega^{n-1} + \cdots + aq_1\Omega + aq_0 = \sum_{i=0}^{n} aq_i \Omega^i \\
aq_n &= a_n^2, \quad aq_0 = a_0^2 \\
aq_i &= a_i^2 - 2a_{i-1}a_{i+1} + 2a_{i-1}a_{i+1} + \cdots \\
&= a_i^2 + 2\sum_{i=j}^{m}(-1)^n a_{i-1}a_{i+1}, \quad m = \min(i, n-i) \\
&= \mu a_i^2 \\
\mu &= 1 + 22\sum_{i=j}^{m}\frac{(-1)^n}{\gamma_{ij}}
\end{aligned} \quad (27)$$

In this way the coefficient $aq_i$ of $PP_{(\Omega)}$ is expressed by the coefficient $a_i$ and the stability index of high order $\gamma_{ij}$, which, in turn, is expressed by stability index $\gamma_i$.

**5.3 Implementation of s-CDM for Hover Control**
To implement s-CDM for the hover control the plant polynomial equations are rewritten as:

$$\begin{aligned} s^{nc} u &= u_{nc} \\ A_p(s) y &= B_p(s) u \end{aligned} \quad (28)$$

The LQR design for hover has the goal to minimize any deviation from hover trim condition with minimum control effort. It is formulated as the following cost function:

$$J = \int_0^\infty [\sum_{i=0}^{nc} qu_i u_i^2 + \sum_{i=0}^{np-1} qy_i y_i^2] dt \quad (29)$$

where $qu_i's$ and $qy_i's$ are scalar constants and $np$ is the order of $A_p(s)$. The weight of the tracking performance $Q$ is expressed as

$$Q = diag([qu_{nc-1}...qu_0 \ qu_{np-1}...qy_1 \ qy_0]) \quad (30)$$

The weight for the control $R$ can then be determined by considering the trade-off of tracking performance and control minimization.

The result of formulation for calculating the weight matrix Q is as follows

$$\begin{aligned}
PP(\Omega) &= Q_u(\Omega) AA_p(\Omega) + Q_y(\Omega) BB_p(\Omega) \\
PP(\Omega) &= P(-s)P(s) = \sum_{i=0}^{nc+np} aq_i \Omega^i \\
AA_p(\Omega) &= A_p(-s)A_p(s) = \sum_{i=0}^{np} apq_i \Omega^i \\
BB_p(\Omega) &= B_p(-s)B_p(s) = \sum_{i=0}^{mp} bq_i \Omega^i \\
Q_u(\Omega) &= \sum_{i=0}^{nc} qu_i \Omega^i \\
Q_y(\Omega) &= \sum_{i=0}^{np-1} qy_i \Omega^i
\end{aligned}$$

Here $mp$ is the order of the plant numerator polynomial $B_p(s)$. In this approach, if $PP(\Omega)$ is obtained as the result of CDM design, the weight polynomials $Q_u(\Omega)$ and $Q_y(\Omega)$ can be obtained.

The calculated results for control design for hovering X-Cell 60 SE are presented as follows.

The helicopter's characteristic polynomials for are expressed for the longitudinal mode and lateral directional mode as follows:

$$\Delta_{LongVer(s)} = s^5 + 31.6547s^4 + 321.7496s^3 + 41.4357s^2 + 11.0203s + 0.9581$$

$$\Delta_{LatDir(s)} = s^5 + 50.7248s^4 + 603.6828s^3 + 42.5467s^2 - 17.8504s + 0.6663$$

(31)

The designed closed-loop system's characteristic polynomials (CDM design) are also given for both modes as:

$$P_{LongVer(s)} = s^6 + 26.6667s^5 + 355.5556s^4 + 2370.37s^3 + 7901.235s^2 \cdots$$
$$\cdots + 13168.72s + 1211.08$$

(32)

$$P_{LatDir(s)} = s^6 + 20s^5 + 200s^4 + 1000s^3 + 2500s^2 + 3125s - 0.3108$$

(33)

In s-CDM framework, the expressions for the plant's characteristic square polynomials are:

$$AP_{LongVer(\Omega)} = \Omega^5 + 358.5208\Omega^4 + 100921.5997\Omega^3 - 5313.9622\Omega^2 \cdots$$
$$\cdots + 42.0507\Omega + 0.9179$$

(34)

$$AP_{LatDir(\Omega)} = \Omega^5 + 1365.6397\Omega^4 + 360080.9158\Omega^3 + 23429.7538\Omega^2 \cdots$$
$$\cdots + 261.9366\Omega + 0.4440$$

(35)

Meanwhile, we have the expressions for the designed closed-loop system's characteristic square polynomials:

$$PP_{LongVer(\Omega)} = \Omega^6 + 15802.4691\Omega^4 + 684773.6626\Omega^3 + 6242950.7697\Omega^2 \cdots$$
$$\cdots + 34683059\Omega + 77073466.2926$$

(36)

$$PP_{LatDir(\Omega)} = \Omega^6 + 5000\Omega^4 + 121875\Omega^3 + 625000\Omega^2 + 1953125\Omega + 2441406.25$$

(37)

Plugging in the above polynomials into Eq.(31), we can obtain the weight of the CDM-LQ design. The weight matrices for the longitudinal and lateral directional mode are given as follows:

$$Q_{LongVer} = \begin{bmatrix} 237.4176 & 0 & 0 & 0 & 0 \\ 0 & 24650646.8931 & 0 & 0 & 0 \\ 0 & 0 & 4981280.8667 & 0 & 0 \\ 0 & 0 & 0 & 34693042.4976 & 0 \\ 0 & 0 & 0 & 0 & 77073684.2119 \end{bmatrix}$$

$$Q_{LatDir} = \begin{bmatrix} 260.0106 & 0 & 0 & 0 & 0 \\ 0 & 93723302.4926 & 0 & 0 & 0 \\ 0 & 0 & 6716721.8342 & 0 & 0 \\ 0 & 0 & 0 & 2021230.8532 & 0 \\ 0 & 0 & 0 & 0 & 2441521.6901 \end{bmatrix}$$

The corresponding CDM-LQ gain matrices are calculated as the following:

$$K_{LongVer} = \begin{bmatrix} -6.04073 \times 10^{-7} & -0.56477 & 2.4429 \times 10^{-4} & 5.2488 \times 10^{-4} & -3.4474 \times 10^{-4} \\ -5.3387 \times 10^{-4} & 5.4885 \times 10^{-4} & 0.23486 & 0.67320 & 0.17247 \end{bmatrix}$$

$$K_{LatDir} = \begin{bmatrix} -9.5475 \times 10^{-7} & -0.99923 & 2.1331 \times 10^{-4} & 4.0244 \times 10^{-6} & -1.1787 \times 10^{-4} \\ -1.114 \times 10^{-3} & 7.9840 \times 10^{-4} & 0.24553 & 0.16685 & 0.10143 \end{bmatrix}$$

Finally, the result of the design is presented in the following figures:

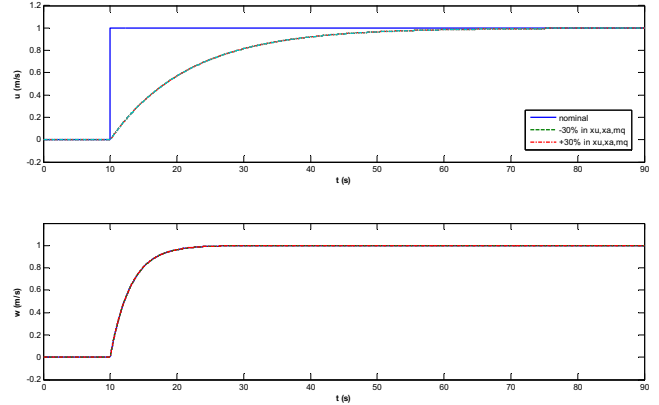

Figure 10 Step response subjected to parameter variation

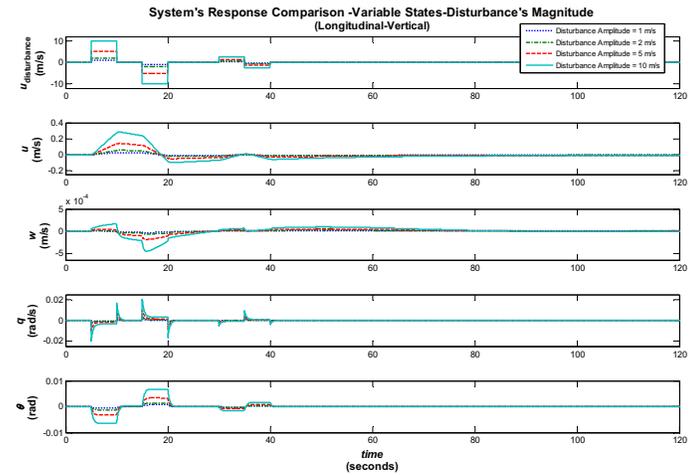

Figure 11 System response comparison for different magnitude of disturbance (longitudinal mode)

## 6  Conclusions

The paper presents the formulation of the control for small-scale helicopter in the algebraic approach framework. In this approach the characteristic polynomial and the controller are design simultaneously with due consideration to the performance specification and constraint imposed to the controller. In CDM, the performance specification is rewritten in a few parameters (stability indices $\gamma_i's$, the stability limit indices $\gamma_i^*{'s}$ and the equivalent time constant, $\tau$ ). These design parameters determine the coefficients of the characteristic polynomial which are related to the controller parameters algebraically in explicit form. The control of small scale helicopter are designed by CDM is shown to be robust against model parameter uncertainties and external disturbances

The elaboration in the paper includes the formal implementation of CDM for aerospace MIMO problem by using LQR framework. In the proposed framework, the considerations for stability, robustness and optimality are addressed simultaneously for a MIMO problem. Beyond the design examples available in the literature that are limited to SISO and SIMO applications, the work demonstrates a successful implementation of CDM-based LQR technique without the need of decomposing a MIMO problem into a series of SISO or SIMO problems.

Finally the extension of the use of Squared Coefficient Diagram Method (s-CDM) for MIMO problem is presented. To the best of author's knowledge, to date the application of s-CDM is limited to a simple SIMO problem. The work proposed the use of s-CDM in conjunction with LQ design in which analytical weight selection is presented. The designed control using s-CDM approach is demonstrated for hovering control of small-scale helicopter simultaneously subjected to plant parameter uncertainties and wind disturbances.